\documentclass[letterpaper]{article} 
\usepackage{aaai24}  
\usepackage{times}  
\usepackage{helvet}  
\usepackage{courier}  
\usepackage[hyphens]{url}  
\usepackage{graphicx} 
\usepackage{natbib}  
\usepackage{caption} 
\usepackage{algorithm}
\usepackage{algorithmic}
\usepackage{amssymb}
\usepackage{amsmath}
\usepackage{multirow}
\usepackage{subcaption}
\usepackage{newfloat}
\usepackage{listings}
\DeclareCaptionStyle{ruled}{labelfont=normalfont,labelsep=colon,strut=off} 
\floatstyle{ruled}
\newfloat{listing}{tb}{lst}{}
\floatname{listing}{Listing}
\title{Learning Invariant Inter-pixel Correlations for
Superpixel Generation}
\author{
    Sen Xu\textsuperscript{\rm 1,2},
    Shikui Wei\thanks{Corresponding author}\textsuperscript{\rm 1,2},
    Tao Ruan\textsuperscript{\rm 3},
    Lixin Liao\textsuperscript{\rm 4}
}
\affiliations{
    \textsuperscript{\rm 1}Institute of Information Science, Beijing Jiaotong University \\
    \textsuperscript{\rm 2}Beijing Key Laboratory of Advanced Information Science and Network Technology\\
    \textsuperscript{\rm 3}Frontiers Science Center for Smart High-speed Railway 
System, Beijing Jiaotong University \\
    \textsuperscript{\rm 4} DaoAI Robotics Inc.\\
    \{senxu1, shkwei, ruantao\} @bjtu.edu.cn, liaolixin@163.com
}

\usepackage{bibentry}

\begin{document}

\maketitle

\begin{abstract}
Deep superpixel algorithms have made remarkable strides by substituting hand-crafted features with learnable ones. Nevertheless, we observe that existing deep superpixel methods, serving as mid-level representation operations, remain sensitive to the statistical properties (e.g., color distribution, high-level semantics) embedded within the training dataset. Consequently, learnable features exhibit constrained discriminative capability, resulting in unsatisfactory pixel grouping performance, particularly in untrainable application scenarios. To address this issue, we propose the \textbf{C}ontent \textbf{D}isentangle \textbf{S}uperpixel (\textbf{CDS}) algorithm to selectively separate the invariant inter-pixel correlations and statistical properties, i.e., style noise. Specifically, We first construct auxiliary modalities that are homologous to the original RGB image but have substantial stylistic variations. Then, driven by mutual information, we propose the local-grid correlation alignment across modalities to reduce the distribution discrepancy of adaptively selected features and learn invariant inter-pixel correlations. Afterwards, we perform global-style mutual information minimization to enforce the separation of invariant content and train data styles. The experimental results on four benchmark datasets demonstrate the superiority of our approach to existing state-of-the-art methods, regarding boundary adherence, generalization, and efficiency. Code and pre-trained model are available at https://github.com/rookiie/CDSpixel.

\end{abstract}

\section{Introduction}

Superpixel segmentation \cite{achanta2012slic,liu2011ERS,achanta2017SNIC,yang2020superpixel-FCN,wang2021ainet} divides an image into compact and contiguous regions of pixels based on specific criteria such as color similarity, texture, or brightness. Compared to pixel-level processing, superpixels significantly reduce the number of image primitives and preserve the structural information, thus improving the efficiency and accuracy of many downstream tasks, i.e., semantic segmentation \cite{he2015supercnn,zhu2014saliency,kwak2017weakly,gadde2016BI}, stereo matching \cite{yang2020superpixel-FCN,birchfield1999multiway,wang2008region}, self-supervised pretraining \cite{sp_pretrain}, image classification \cite{zhao2022sp-classification, guo2018fuzzy}, etc. The usual practice of traditional superpixel algorithms is to initialize a regular grid of superpixels and iteratively adjust the association of pixels and neighboring superpixels. Afterward, deep learning algorithms leverage image features learned by neural networks to replace hand-crafted features, i.e., XYlab used by traditional methods, significantly improving the performance of superpixel algorithms. 

\begin{figure}
    \centering
    \includegraphics[width=\columnwidth]{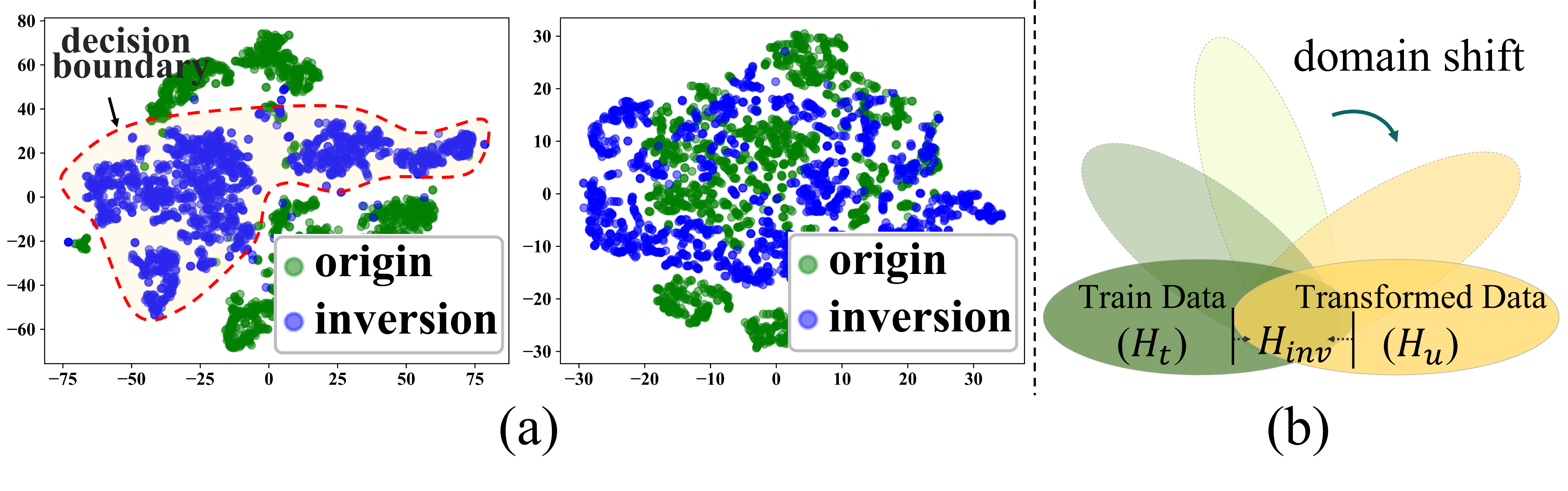}
    \caption{Motivation. (a) Visualization of the t-SNE distributions on the BSDS dataset. From left to right are the baseline and our CDS. After applying color inversion, the feature distribution of baseline displays a noticeable decision boundary. In contrast, the feature distribution extracted by CDS is more compact (i.e.,[-80, +80] vs [-40, +40]) and indivisible. (b) Gradually modifying the stylistic information of both auxiliary and original data enhances the purity of the shared invariant inter-pixel correlations.}
    \label{fig:t-sne}
\end{figure}

However, although learnable features have been proven effective, they also introduce new challenges. Superpixel segmentation, as a mid-level image representation task, needs to adapt to a variety of open-world scenarios. Traditional superpixel algorithms employ independent online processes (e.g., clustering or graph-cut), which remain unaffected by inter-instance influences. In contrast, deep superpixel algorithms carry the potential risk of learning the unique data distribution present in the training set. To validate this potential risk, we conducted a straightforward experimental verification in Fig. \ref{fig:t-sne}(a). Given that this concern reflects the algorithm's generalization ability, we opted for SCN \cite{yang2020superpixel-FCN}, a deep superpixel algorithm renowned for its solid generalization, for experimental validation. We applied an inversion operation (subtracting each pixel value from 255) to simulate variations in the data distribution within the test set. While the inter-pixel correlations of identical images should remain constant through the linear transformation of RGB values during superpixel generation, the features extracted by the baseline algorithm exhibit a discernible decision boundary. Hence, learnable superpixels not only encompass a pure representation of inter-pixel correlations, but are also influenced by stylistic noise from the training set, such as high-level semantics and color distribution. Among the previous works, this issue is overlooked even though necessary. 

To address this issue, we introduce a novel deep superpixel method named the Content Disentangle Superpixel (CDS) algorithm to disentangle the dataset-specific style from the invariant content, i.e., the inter-pixel correlation used for superpixel segmentation. Our CDS achieve the objective of decoupling by constructing auxiliary data. As superpixel segmentation focuses not on high-level semantics but on pixel correlations, auxiliary data, despite altering style, do not disrupt the inherent pixel relationships. As illustrated in Fig.\ref{fig:t-sne}(b), gradually modifying the stylistic information of both auxiliary and original data enhances the purity of the shared invariant inter-pixel correlations. Concretely, our method consists of three parts in order: feature extraction, content disentangle (CD), and superpixel generation. In the CD phrase, we feed the modality embeddings into the content selective gate to  adaptively decouple content features and style noise. Afterwards, we propose the superpixel-grid correlation alignment to ensure that the selected content features from two modalities have the same pixel correlations. To prevent the occurrence of degenerate solutions in the content selective gate above. We enforce constraints to encourage the dataset/modal-specific features to have smaller mutual information while avoiding degradation of the selected modality style.  Finally, a modality-shared superpixel decoder is designed to predict superpixel associations. Moreover, the auxiliary modality is only used in the training phase. In summary, our contributions are: 
\begin{itemize}
    \item We discover that existing deep superpixel algorithms depend on the distribution of training data and propose the CDS algorithm to ensure that learnable superpixels have a high generalization and boundary adherence power.
    \item During the training phase, CDS introduces an auxiliary modality to help decouple the correlated features among pure pixels in the RGB modality; while performing inference using only RGB. Compared to previous work, our algorithm does not increase additional computational burden but effective.
    \item Experimental results on datasets from four different domains indicate the superiority of our approach to existing superpixel algorithms. In addition, we demonstrate that our method improves the performance of downstream tasks.
\end{itemize}

\section{Related Work}
\label{Sec:related}
\textbf{Traditional Superpixel Algorithms.} The research of traditional superpixel algorithms \cite{achanta2012slic,bergh2012seeds,felzenszwalb2004FH,li2015LSC,liu2011ERS,grady2006RW,watershed} has a long timeline. In times of scarce parallel computing resources, traditional superpixel algorithms can reduce image information redundancy and are often used to improve the computational efficiency of downstream tasks. Traditional superpixel algorithms are mainly classified into graph-based, clustering-based, and energy-based approaches. Concretely, graph-based methods, e.g., ERS \cite{liu2011ERS}, treat superpixel segmentation as a graph partitioning problem. The image is formulated as an undirected graph, and the edge weight indicates the inter-pixel similarity. Clustering-based methods, e.g., SLIC \cite{achanta2012slic}, SNIC \cite{achanta2017SNIC}, and LSC \cite{li2015LSC}, initial the superpixels with seed pixels and apply cluster algorithms like k-means to adjust the pixel association. Energy-based methods, e.g., SEEDS \cite{bergh2012seeds} and ETPS \cite{yao2015ETPS}, first partition the image into regular grids and  leverage different energy functions as an objective to exchange the pixels between neighboring superpixels. Traditional superpixel algorithms usually use CIELAB colors, concatenated with two-dimensional position encoding as pixel features.

\noindent\textbf{Deep Superpixel Algorithms.} SEAL \cite{SEAL} combines the neural networks with the ERS \cite{liu2011ERS} by proposing the segmentation-aware affinity loss to learn cluster-friendly features. SEAL is not end-to-end trainable. To address this problem, Jampani \textit{et al.} relaxes the nearest neighbor constraints of SLIC and develops the first differentiable deep algorithm SSN \cite{SSN}. Since both SEAL and SSN require iterative traditional superpixel operations to complete the segmentation, SCN \cite{yang2020superpixel-FCN} model the superpixel segmentation as a classification problem between each pixel and its neighboring nine superpixels, which significantly improves the computational efficiency of superpixel segmentation. Based on the SCN, Wang \textit{et al.} \cite{wang2021ainet} propose the association implantation module, i.e., AINet, to enable the network to enhance the relations between the pixel and its surrounding grids. Among them, SCN and AINet are the SOTA algorithms with optimal performance.

\noindent\textbf{Style Removal Algorithms.} To better understand our work, we introduce the style removal techniques related to our Motivation. Style removal is not an independent visual task, often used as a technical means to solve problems such as style transfer and domain generalization. Existing methods mainly focus on two aspects: Normalization \cite{pan2018INtwo, ulyanov2017IN, huang2017adaIN} and Whitening \cite{li2017whiten, pan2019s_whitening,cho2019GIW, choi2021robustnet}. Especially in \cite{ulyanov2017IN}, the authors propose instance normalization to prevent overfitting on the domain-specific style of training data. \cite{pan2018INtwo} achieve significant performance improvement by incorporating the IN layers to capture style-invariant information. \cite{li2017whiten} visually validates that the image style exists mainly on the correlation of feature channels and proposes the whitening transform to extract image content. Channel whitening calculates the covariance matrix across all channels, resulting in excessive computational complexity ($O(n^2)$). In \cite{cho2019GIW}, the authors introduce group-wise instance whitening (GIW) transform to improve time efficiency. Overall, instance-level normalization and whitening both ensure that feature channels are independent of each other.

\section{Preliminaries}
Superpixel segmentation is the task of partitioning an input image $\mathcal{I}\in \mathbb{R}^{H\times W \times3}$ into a set of n superpixels $\mathcal{S} = \{S_1, S_2, ..., S_n\}$, where each superpixel $S_i$ consists of a group of adjacent pixels with similar characteristics. Mathematically, superpixel segmentation algorithms aim to calculate the pixel-superpixel association map $\mathcal{Q}$. Since computing the association between each pixel $p$ with n superpixels has a high computational complexity, the latest deep superpixel approaches \cite{yang2020superpixel-FCN} formulate the superpixel segmentation as a local classification problem between pixel p with nine neighbor superpixel grids, and  improve the time efficiency. This also served as our theoretical foundation. 

Concretely, the image $\mathcal{I}$ is initialized into regular grids, and $\mathcal{Q}\in \mathbb{R}^{H\times W \times 9}$ indicates the probability that each pixel $p$ is attributed to its nine nearby superpixel grids. The association map $\mathcal{Q}$ is predicted by a series of CNNs. Since no label is available for this output, deep algorithms design the superpixel loss inspired by the k-means convergence condition. Formally, let $f(p)$ be the pixel properties, i.e., semantic one-hot vector and positional encoding; we first obtain the superpixel clustering center properties $g(s)$ leveraging $\mathcal{Q}$ and $f(p)$. Then reconstruct the pixel property as follows:

\begin{equation}
g(s)=\frac{\sum_{\{p| s \in N_p\}} f(p) \cdot \mathcal{Q}_s(p)}{\sum_{\{p| s \in N_p\}} \mathcal{Q}_s(p)}, f^{\prime}(p)=\sum_{s \in N_p} g(s) \cdot \mathcal{Q}_s(p).
\end{equation}
Here $N_p$ indicates the set of adjacent superpixels of $p$, and the $\mathcal{Q}_s(p)$ is the predicted probability that pixel $p$ is assigned to superpixel $s$. Finally, superpixel segmentation amounts to minimizing the reconstruction distance
\begin{equation}
L_{sp}(Q)= \sum_p \mathrm{Dis} \left(f(p), f^{\prime}(p)\right).
\end{equation}
Following previous works \cite{yang2020superpixel-FCN,wang2021ainet}, we use the cross-entropy and the Euclidean distance as
the distance measure of the semantic label and the position vector, respectively.

\begin{figure}
    \centering
    \includegraphics[width=\columnwidth]{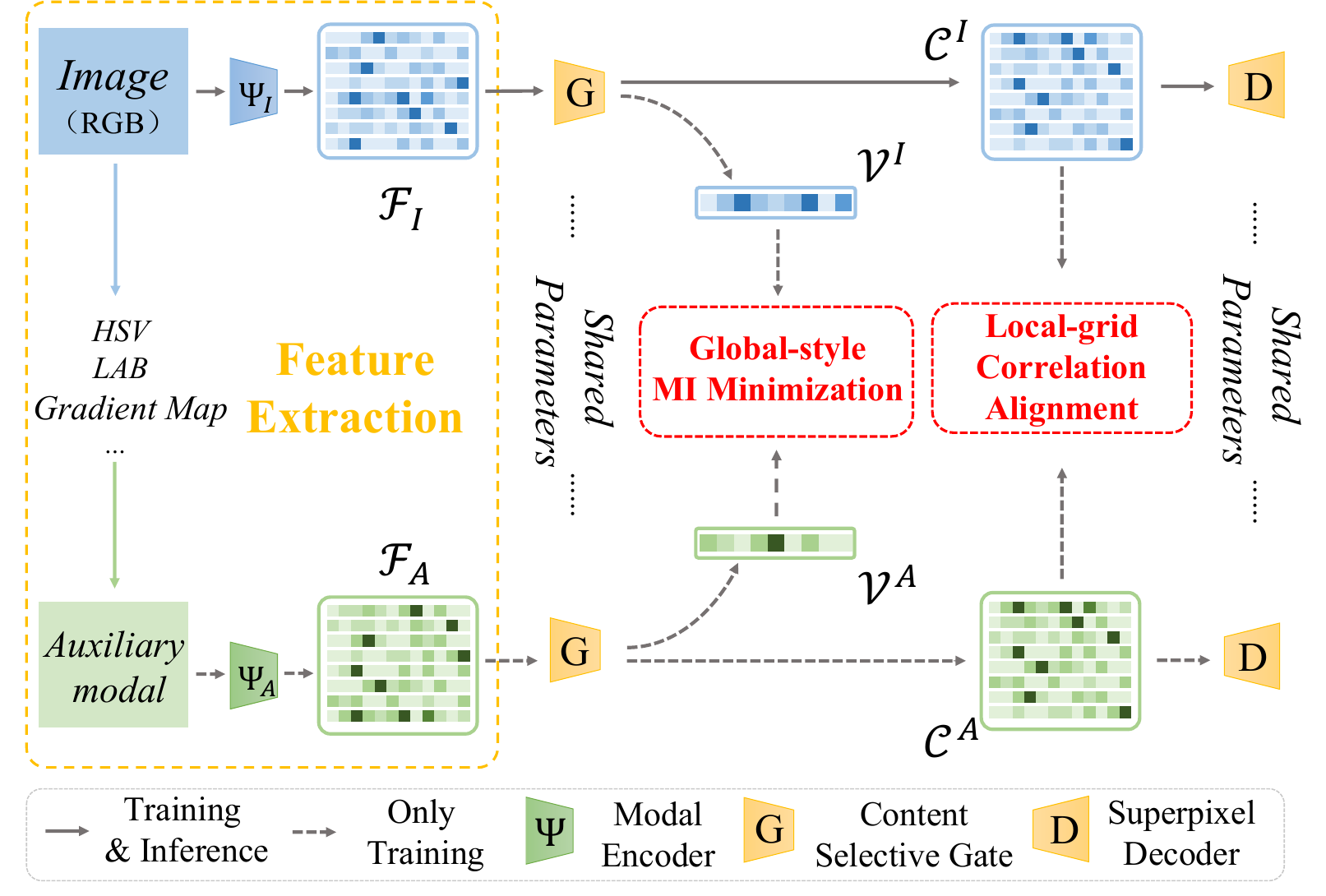}
    \caption{Flowchart of the proposed content disentangle superpixel algorithm.}
    \label{fig:flowchart}
    \vspace{-1em}
\end{figure}

\section{Proposed Algorithm}

In this section, we present Content Disentangle Superpixel (CDS) algorithm (shown in Fig.\ref{fig:flowchart}), which aims to learn invariant inter-pixel correlations and reduce the style noise for superpixel generation. We first introduce the auxiliary modal and feature extraction, and then, present the content disentangle mechanism. Finally, the superpixel decoder is shown in the last subsection.

\subsection{Auxiliary Modal and Feature Extraction}

\textbf{Auxiliary Modal.} As shown in Fig.\ref{fig:t-sne}(b), we construct samples with large domain offsets homogeneously sourced from the raw data to separate the inter-pixel correlation information, and the constructed auxiliary samples need to: \textit{ (1) preserve the pixel interrelations intact, (2) and exhibit significant stylistic differences.} Consequently, we adopt the auxiliary modality. Modality refers to the way in which something happens or is experienced \cite{multimodaldefination}, and different modalities, i.e., HSV and LAB color-space transform and gradient map, describe objects from different perspectives. Although these modalities do not possess as significant modality barriers as heterogeneous data like text and images, they exhibit substantial variations in pixel-level descriptions.

\noindent\textbf{Modal Encoder.} Different from \cite{yang2020superpixel-FCN, wang2021ainet}, to facilitate the decoupling of learnable image features, we remain in the explicit pixel-level feature extraction phase. For each image $\mathcal{I}$ and its auxiliary modality sample $\mathcal{A}\in \mathbb{R}^{H\times W \times 3}$, the whole process can be formulated as:
\begin{equation}
    \Psi(\mathcal{I})=\mathcal{P}\left\{\Phi\left(\mathcal{I}; \theta\right) + \mathcal{I};\theta^{*}\right\}
\label{eq:f1}
\end{equation}
\begin{equation}
    \Psi(\mathcal{A})=\mathcal{P}\left\{\Phi\left(\mathcal{A}; \gamma\right) + \mathcal{A};\gamma^{*}\right\}
\label{eq:f2}
\end{equation}
where the $\Phi$ is a series of convolution layers modified from the CNN part of SSN \cite{SSN}. We adopt its structure for simplicity and efficiency. Since we formulate superpixels as a local classification problem, we do not concatenate positional encoding as SSN does. Additionally, to prevent the loss of pixel information, we additionally use a non-linear mapping function $\mathcal{P}\{\cdot\}$ to aggregate the original pixel values. $\theta, \theta^{*}$ and $\gamma, \gamma^{*}$ indicate the parameters of two modalities, respectively.

\subsection{Content Disentangle Mechanism}

After feature extraction process, we obtain the pixel-level embedding of the inputs, i.e., $\mathcal{F}_{\mathcal{I}}, \mathcal{F}_{\mathcal{A}} \in \mathbb{R}^{C \times H\times W}$. In this section, we introduce the content disentangle mechanism to adaptively select the inter-pixel correlation information, which contains three operations in detail.

\noindent\textbf{Content Selective Gate} is a learnable filter to separate the pixel embedding $\mathcal{F}_{\mathcal{A},\mathcal{I}}$ into content features $\mathcal{C}^{\mathcal{A},\mathcal{I}} \in \mathbb{R}^{C \times H\times W}$ and global modality/dataset style vectors $\mathcal{V}^{\mathcal{A},\mathcal{I}} \in \mathbb{R}^{C}$. As discussed in the related work, the image style exists mainly on the correlation of feature channels \cite{li2017whiten}. More importantly, superpixels do not consider global semantics like classification tasks do, and each position in the spatial domain has equal significance. Therefore, the content selective gate is designed as a share-weighted channel-wise attention strategy:
\begin{equation}
\begin{aligned}
    &\mathcal{C}^{i} =\mathrm{Gate}(\mathcal{F}_{i}) \cdot \mathcal{F}_{i} \\
    &\mathcal{V}^{i}=\mathrm{avgpool}((1-\mathrm{Gate}(\mathcal{F}_{i})) \cdot \mathcal{F}_{i})
\end{aligned}
\label{eq:g1}
\end{equation}
where $i \in \{\mathcal{A},\mathcal{I}\}$, $\mathrm{avgpool}$ is the channel-wise average pooling operation, and 
\begin{equation}
\mathrm{Gate}(\mathcal{F}_{i})=\sigma\left(\mathrm{W}_2 \cdot \delta\left(\mathrm{W}_1 \cdot \mathrm{avgpool}(\mathcal{F}_{i})\right)\right).
\label{eq:g2}
\end{equation}
Here, $\mathrm{W}_1$ and $\mathrm{W}_2$ are the parameters for two fully-connected layers, and the $\delta(\cdot)$ and $\sigma(\cdot)$ are the relu and sigmoid function, respectively. 

\begin{figure}
    \centering
    \includegraphics[width=0.9\columnwidth]{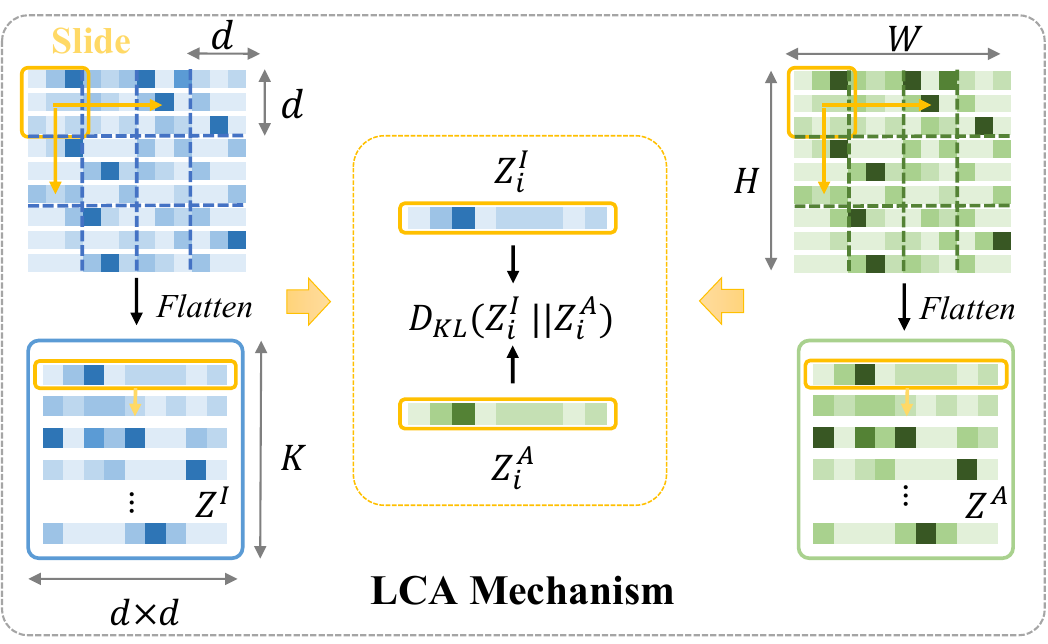}
    \caption{Illustration of the Local-grid Correlation Alignment (LCA) mechanism. LCA performs spatial domain distribution alignment at the superpixel level.}
    \label{fig:align}
     \vspace{-1em}
\end{figure}

\noindent\textbf{Local-grid Correlation Alignment} is proposed to enforce the auxiliary modality and the primary modality to learn similar superpixel-friendly invariant features, i.e., the correlations between pixels. For superpixels, our objective is to ensure that embeddings of neighboring pixels with similar attributes are highly similar, while those of dissimilar pixels exhibit pronounced distinctions. This perspective highlights that superpixels prioritize capturing variations in feature dissimilarity at each image position, rather than being fixated on specific numerical values. Consequently, it becomes essential to guarantee the congruence of spatial distributions between the auxiliary and primary modalities. 

As illustrated in Fig.\ref{fig:align}, given content features $\mathcal{C^{I}},\mathcal{C^{A}} \in \mathbb{R}^{C \times H\times W}$ and initial superpixel grid with $d\times d$ size, We first employ average pooling to reduce dimensions, followed by traversing all superpixel grids and unfolding features in the spatial domain, resulting in the feature matrix $Z\in\mathbb{R}^{K\times d^2}$. $K$ is the initial superpixel number which calculated by $(H/d)\times(W/d)$. Then, the local-grid correlation alignment
constraint is formulated as:
\begin{equation}
\mathcal{L}_{align}=\frac{1}{K}\sum_{i}^K\mathcal{D}_{K L}( Z^I_i  \| Z^A_i ).
\end{equation}

Here, we do not directly compute the global spatial distribution for three main reasons: (1) To enhance parallel computing efficiency. (2) To prevent the loss of local information, as softmax normalization is employed prior to calculating KL divergence, and a large number of pixels could lead to excessively small local probabilities. (3) The deep superpixel algorithm is defined as a neighborhood classification problem.

\begin{algorithm}[t]

\caption{Training pseudocode for CDS.}
\begin{algorithmic}
  \STATE \textbf{Input:} Input image $\mathcal{I}$ and auxiliary modal $\mathcal{A}$.
  \STATE \textbf{Output:} The superpixel association map $\mathcal{Q}_{\mathcal{I}}$ and $\mathcal{Q}_{\mathcal{A}}$.
  \STATE Initialize components: $\Psi_{\mathcal{I}}$, $\Psi_{\mathcal{A}}$, $\operatorname{Gate}$, and superpixel decoder $D$. Initialize the variational distribution network $h^{\theta}$.
  \FOR{each training iteration} 
    \STATE \textbf{Step 1}: Update the main network. (Fix the $h^{\theta}$.)
    \STATE Conduct auxiliary modal $\mathcal{A}$.
    \STATE \quad Calculate the $\mathcal{F}_{\mathcal{I}}$ and $\mathcal{F}_{\mathcal{A}}$ by Eq.\ref{eq:f1}, Eq.\ref{eq:f2}.
    \STATE \quad Calculate the $\mathcal{C}^{i}$ and $\mathcal{V}^{i}$ by Eq.\ref{eq:g1}, Eq.\ref{eq:g2}.
    \STATE \quad Predict the $\mathcal{Q}_{\mathcal{I}}$ and $\mathcal{Q}_{\mathcal{A}}$ by feeding $\mathcal{C}^{i}$ into the superpixel decoder $\mathcal{D}$.
    \STATE \quad Calculate $\mathcal{L}_{align}$, $\mathcal{L}_{MI}$, and $\mathcal{L}_{sp}$, respectively.
    \STATE \quad Update the CDS
    \STATE \textbf{Step 2}: Update the variational distribution network.
    \STATE \quad Detach the value of $\mathcal{V}^{I}$ and $\mathcal{V}^{A}$ from the computational graph.
    \STATE \quad Calculate $\mathcal{L}(\theta)$.
    \STATE \quad Update the parameters of $h^{\theta}$ 
  \ENDFOR
\end{algorithmic}
\label{alg:train}
\end{algorithm}
\noindent\textbf{Global-style Mutual Information Minimization.} Since only alignment lead to information loss or confusion between modalities, to prevent the occurrence of degenerate solutions in the content selective gate above, we enforce constraints to encourage the dataset/modal-specific features $\mathcal{V}$ to have smaller mutual information. Mathematically, given the style vectors $\mathcal{V}^{\mathcal{A},\mathcal{I}}$, \cite{cheng2020club} provide an upper bound on their mutual information:
\begin{figure*}
  \centering
  \begin{subfigure}[b]{0.228\textwidth}
    \begin{subfigure}[b]{\textwidth}
      \includegraphics[width=\textwidth]{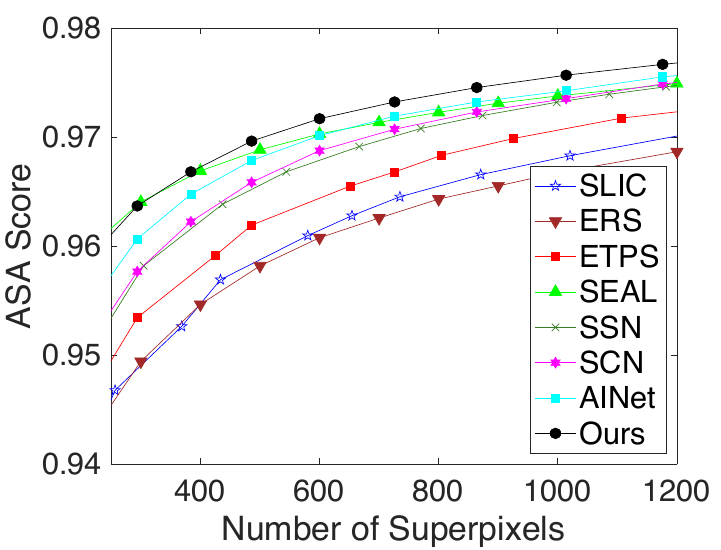}
    \end{subfigure}\\
    \begin{subfigure}[b]{\textwidth}
      \includegraphics[width=\textwidth]{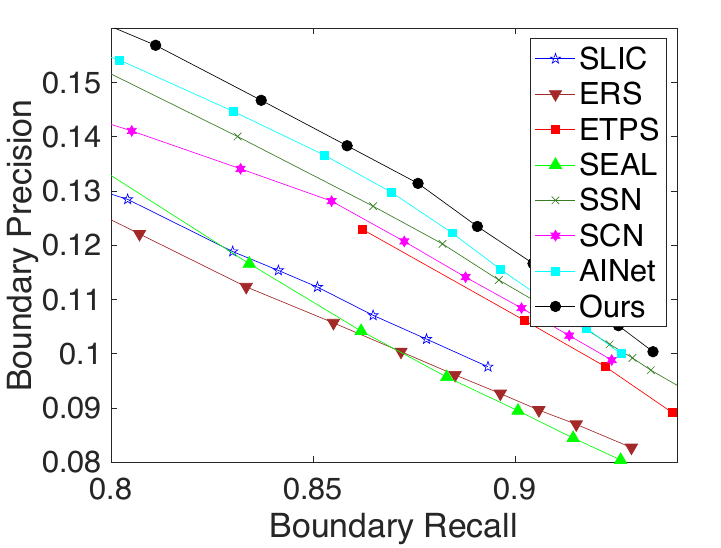}
    \end{subfigure}\\
    \begin{subfigure}[b]{\textwidth}
      \includegraphics[width=\textwidth]{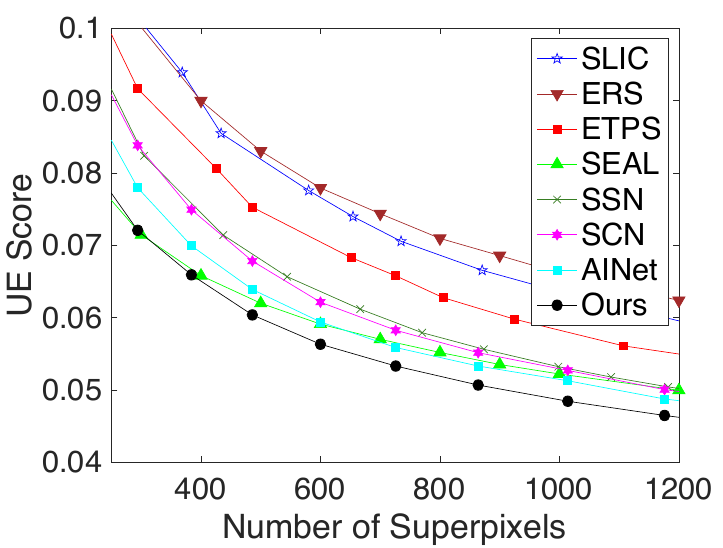}
    \end{subfigure}
    \caption{BSDS500}
  \end{subfigure}
  \begin{subfigure}[b]{0.228\textwidth}
    \begin{subfigure}[b]{\textwidth}
      \includegraphics[width=\textwidth]{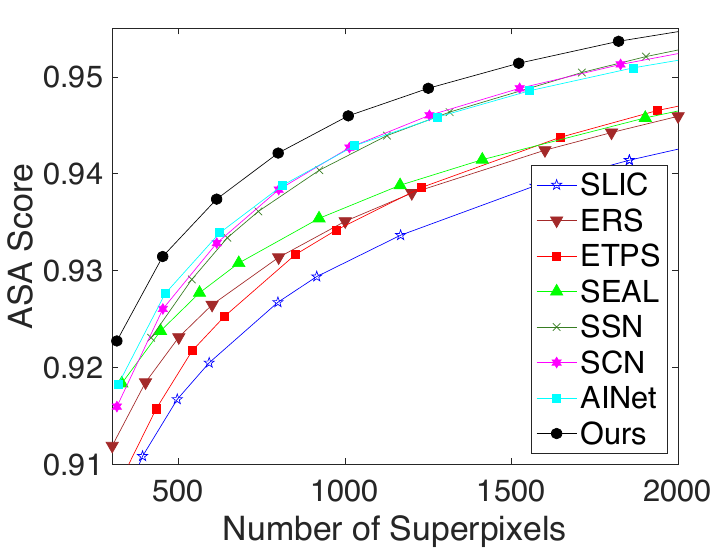}
    \end{subfigure}\\
    \begin{subfigure}[b]{\textwidth}
      \includegraphics[width=\textwidth]{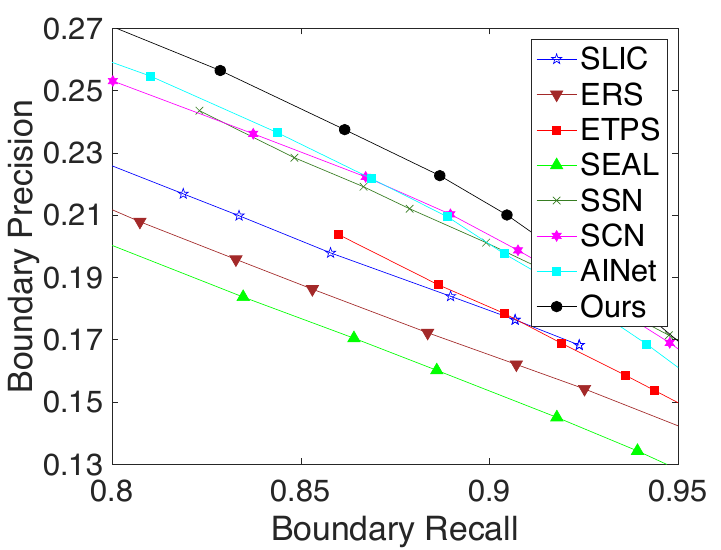}
    \end{subfigure}\\
    \begin{subfigure}[b]{\textwidth}
      \includegraphics[width=\textwidth]{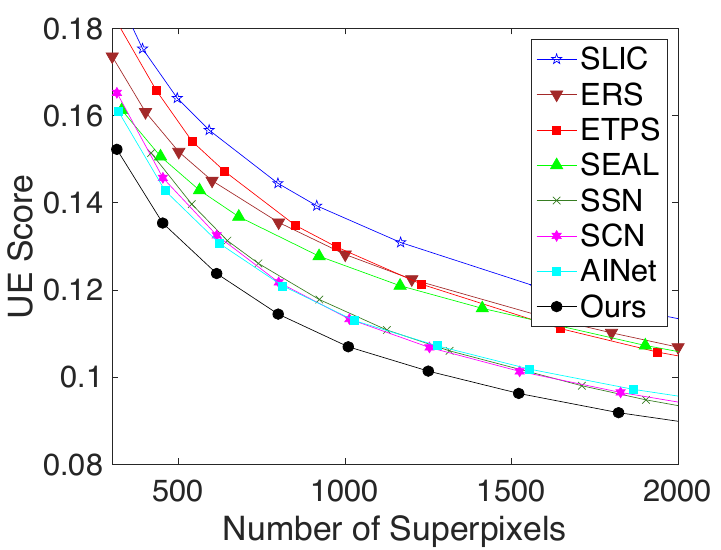}
    \end{subfigure}
    \caption{NYUv2}
  \end{subfigure}
  \begin{subfigure}[b]{0.228\textwidth}
    \begin{subfigure}[b]{\textwidth}
      \includegraphics[width=\textwidth]{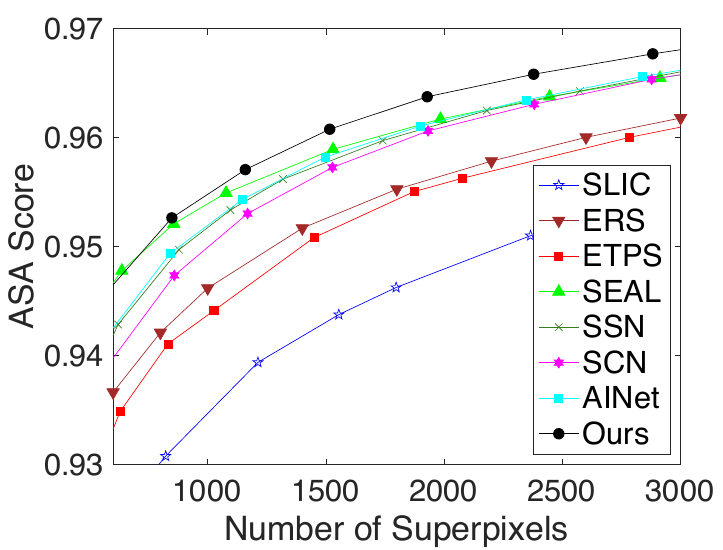}
    \end{subfigure}\\
    \begin{subfigure}[b]{\textwidth}
      \includegraphics[width=\textwidth]{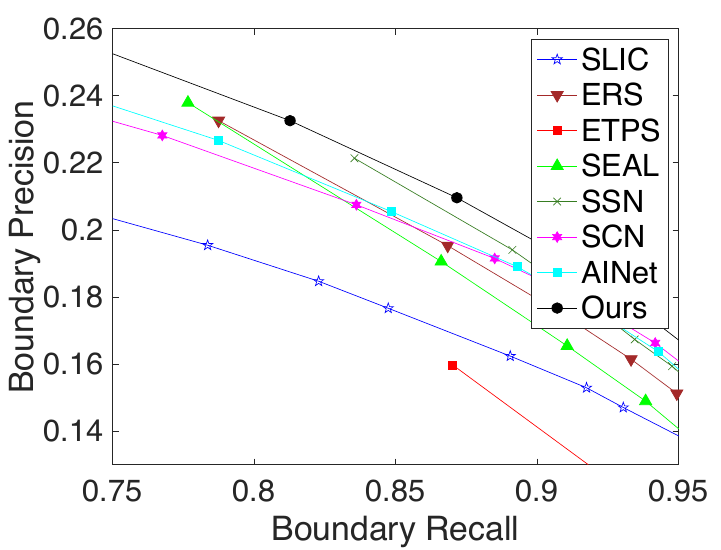}
    \end{subfigure}\\
    \begin{subfigure}[b]{\textwidth}
      \includegraphics[width=\textwidth]{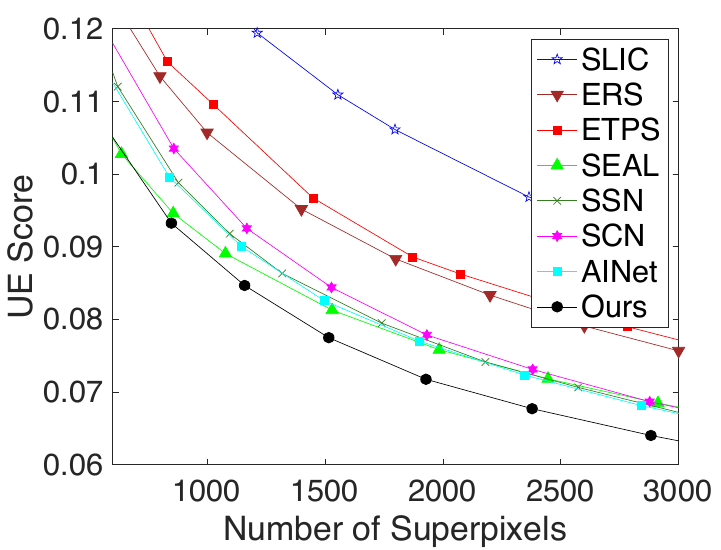}
    \end{subfigure}
    \caption{KITTI}
  \end{subfigure}
  \begin{subfigure}[b]{0.228\textwidth}
    \begin{subfigure}[b]{\textwidth}
      \includegraphics[width=\textwidth]{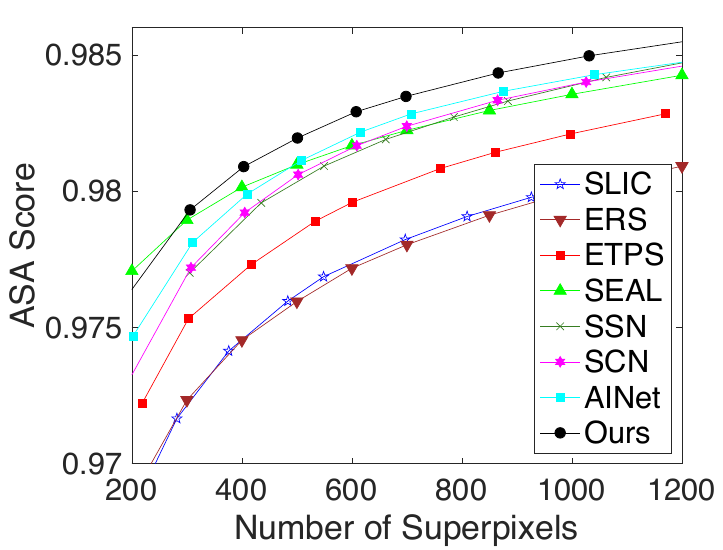}
    \end{subfigure}\\
    \begin{subfigure}[b]{\textwidth}
      \includegraphics[width=\textwidth]{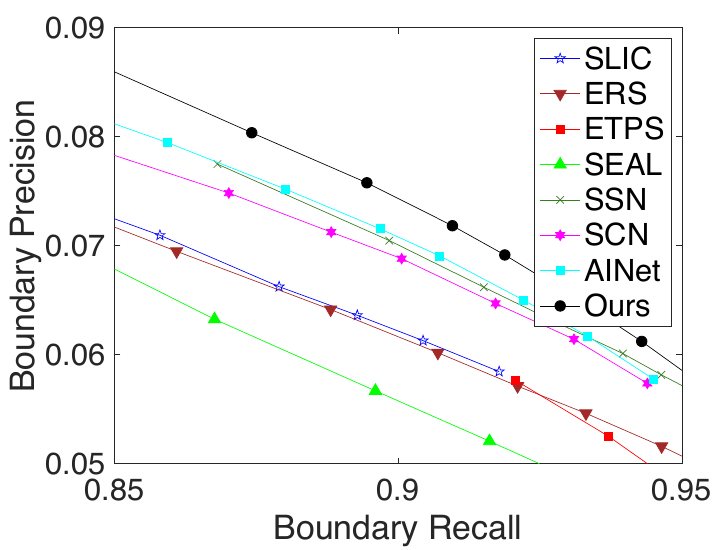}
    \end{subfigure}\\
    \begin{subfigure}[b]{\textwidth}
      \includegraphics[width=\textwidth]{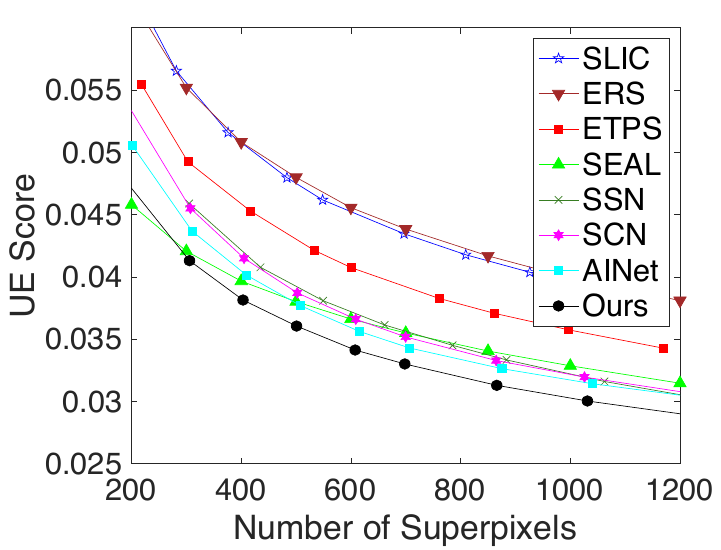}
    \end{subfigure}
    \caption{VOC2012}
  \end{subfigure}
  
  \caption{Performance comparison on four datasets from different domains. From Left to Right: BSDS, NYU, KITTI and VOC datasets. From Top to Bottom: ASA, BR-BP and UE metrics. Except UE, higher values indicate that the algorithm is more effective.}
  \label{fig:results}
\end{figure*}

\begin{equation}
\begin{aligned}
\mathrm{I}(\mathcal{V}^{\mathcal{I}} ; \mathcal{V}^{\mathcal{A}})  \leq &\mathbb{E}_{p(\mathcal{V}^{\mathcal{A}} , \mathcal{V}^{\mathcal{I}})}[\log p(\mathcal{V}^{\mathcal{I}} \mid \mathcal{V}^{\mathcal{A}} )] \\
& -\mathbb{E}_{p(\mathcal{V}^{\mathcal{A}} )} \mathbb{E}_{p(\mathcal{V}^{\mathcal{I}})}[\log p(\mathcal{V}^{\mathcal{I}} \mid \mathcal{V}^{\mathcal{A}} )]
\end{aligned}
\end{equation}
where $\mathrm{I}(\mathcal{V}^{\mathcal{I}} ; \mathcal{V}^{\mathcal{A}}) $ indicates the mutual information. However, the $\mathcal{V}^{\mathcal{A},\mathcal{I}}$ are learnable variables, and conditional distribution $p(\mathcal{V}^{\mathcal{I}} \mid \mathcal{V}^{\mathcal{A}}) $ is unavailable. Consequently, we leverage the variational distribution $h^\theta$ to approximate $p(\cdot| \cdot)$. Then, the proposed MI loss is given to minimize the MI upper bound:

\begin{equation}
    \mathcal{L}_{MI} = \frac{1}{N^2}\sum_{i=1}^{N}\sum_{j=1}^{N}[(\log h^{\theta}(\mathcal{V}^{\mathcal{I}}_i|\mathcal{V}^{\mathcal{A}}_i)) - (\log h^{\theta}(\mathcal{V}^{\mathcal{I}}_j|\mathcal{V}^{\mathcal{A}}_i))] 
\end{equation}

Specifically, the variational distribution $h^\theta(\mathcal{V}^{\mathcal{I}}|\mathcal{V}^{\mathcal{A}}))$ is usually implemented with neural networks and optimized independently by $\mathcal{L}(\theta) = -\frac{1}{N}\sum_{i=1}^{N}h^{\theta}(\mathcal{V}^{\mathcal{I}}_i|\mathcal{V}^{\mathcal{A}}_i)$, the log-likelihood function \cite{yue2022dare}. When participating in the computation of $\mathcal{L}_{MI}$ and updating of the main network, the parameters of $h^{\theta}$ are frozen, and only gradients are passed.

\subsection{Superpixel Generation}

Since the features encoding local pixel correlations, i.e., $\mathcal{C}^{\mathcal{I},\mathcal{A}}$ have the same size as the original image, we initially formulate hierarchical features $\{c_1, c_2, ..., c_n\}$, where $n=\log_2(d)$, employing bilinear interpolation. The size of $c_i$ is $\frac{1}{2^{i-1}}\cdot(H,W), i\in[1,n]$. Then, similar to U-net, we construct the decoder in a bottom-up manner, and the predict association map $\mathcal{Q}$ is given as:

\begin{equation}
    \mathcal{Q}_i = D(c_1,c_2,...,c_n; \epsilon), \quad \forall i \in \{\mathcal{A},\mathcal{I}\},
\end{equation} where $\epsilon$ indicates the parameters of D. In this way, we extend the network's receptive field to encompass the adjacent vicinity of nine $d\times d$ superpixel grids, thereby enabling a more nuanced prediction of the relationship map. 
\begin{equation}
\mathcal{L}_{total}= \mathcal{L}_{align} + \mathcal{L}_{MI} + \mathcal{L}_{sp}(\mathcal{Q}_{\mathcal{I}})+\mathcal{L}_{sp}(\mathcal{Q}_{\mathcal{A}})
\label{Eq:ltotal}
\end{equation}

Finally, the overall optimization objective of the proposed CDS is defined in Eq. \ref{Eq:ltotal}, which comprises three components: the loss for the proposed local-grid correlation alignment, the mutual information loss, and the superpixel loss. Algorithm \ref{alg:train} demonstrates the pseudocode for training our algorithm.

\section{Experiments}

\subsection{Experiment Settings}
\label{sec:dataset}
\textbf{Dataset.} Since the proposed CDS superpixel algorithm is introduced to mitigate the influence of attribute noise from the training set. We evaluate our method on four segmentation datasets from different domains: BSDS500 \cite{bsds500}, NYUv2 \cite{nyuv2}, KITTI \cite{geiger2012kitti}, Pascal VOC2012 \cite{everingham2015voc}. Concretely, BSDS is an object edge detection dataset that contains five segmentation labels from different annotators for each image. NYUv2 is a semantic segmentation dataset for indoor scenes. KITTI is a commonly used street scene segmentation dataset for autonomous driving. Pascal VOC is a benchmark segmentation dataset with twenty object categories, which can be used for tasks such as object detection, instance segmentation.

\begin{table}
  \centering
  \resizebox{0.8\columnwidth}{!}{%
  \begin{tabular}{l|c|c}
    \hline
    \hline
   Model   & Time(ms) &  Device \\ \hline
		SLIC\cite{achanta2012slic}& 120 &  CPU \\ 
		ERS\cite{liu2011ERS}  & 940&  CPU  \\ 
		ETPS\cite{yao2015ETPS} & 82&  CPU  \\
		\hline
		SEAL\cite{SEAL}& 2658 &  GPU\&CPU  \\
		SSN\cite{SSN}  & 278 &   GPU  \\
		SCN\cite{yang2020superpixel-FCN} & 5  &  GPU  \\ 
        AINet\cite{wang2021ainet} & 29 & GPU \\ \hline
		Ours   & 6 &  GPU  \\\hline
    \hline
  \end{tabular}%
  }
  \caption{Runtime comparison for generating about 600 superpixels on NYUv2 with image size $608\times448$.}
  \label{tab:e}
\end{table}

\noindent\textbf{Evaluation protocol.} Follow the evaluation protocol of previous works \cite{yang2020superpixel-FCN, wang2021ainet}, we only train our model on the BSDS500 dataset and run inference on the other datasets. The clustering performance is mainly evaluated by three public metrics: Achievable Segmentation Accuracy (ASA), Boundary Recall-Precision (BR-BP) curve, Under-segmentation Error (UE). 

\noindent\textbf{Implementation details.} During the training phase, we apply data augmentation through random resize, random cropping to $208\times208$, and random horizontal/vertical flipping for our CDS. We trained the models using Adam optimizer. The learning rate starts at 5e-4 and is updated by the poly learning rate policy. We trained our model for 150k iterations with batch size eight, and superpixel grid size $d$ is set to 16. We use the gradient map as the auxiliary modality and conduct all the experiments on single RTX3090 GPU.

\subsection{Comparison with the State-of-the-Arts}
\label{SOTA}

Fig.\ref{fig:results} and Tab.\ref{tab:e} report the metrics and runtime comparisons with representative superpixel algorithms, respectively, including three traditional approaches, i.e., clustering-based method SLIC \cite{achanta2012slic}, graph-based method ERS \cite{liu2011ERS}, and energy-based method ETPS \cite{yao2015ETPS}, and four deep superpixel approaches, i.e., SEAL \cite{SEAL}, SSN \cite{SSN}, SCN \cite{yang2020superpixel-FCN} and AINet \cite{wang2021ainet}. For traditional superpixels, we leverage the hyperparameters posted by \cite{stutz2018superpixels}. For deep superpixel algorithms, we conduct the experiment with their official implementations. Among them, SCN and AINet are the SOTA algorithms with optimal performance.

For a fair comparison, we employ the actual number of generated superpixels due to potential variations from manually specified counts. Our observations reveal the following:
\begin{itemize}
    \item (1) Our method consistently outperforms others across all datasets, with a widening and stabilizing lead as superpixel count increases. This implies that style noise reduction enhances contour accuracy.
    \item (2) Across diverse domains, our superiority becomes more pronounced, highlighting superior generalization. 
    \item (3) The auxiliary modality only participates in computations during the training phase, ensuring both improved algorithm performance and preserved inference speed.
\end{itemize}

\begin{figure*}
  \centering
  \begin{subfigure}[b]{0.11\textwidth}
    \includegraphics[width=\textwidth]{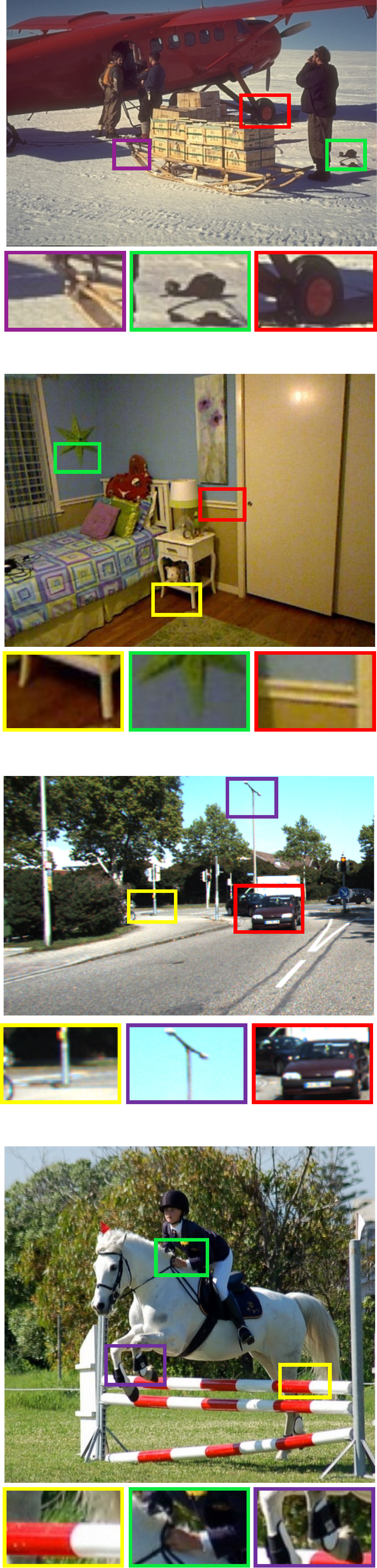}
    \caption{Image}
  \end{subfigure}
  \hfill
  \begin{subfigure}[b]{0.11\textwidth}
    \includegraphics[width=\textwidth]{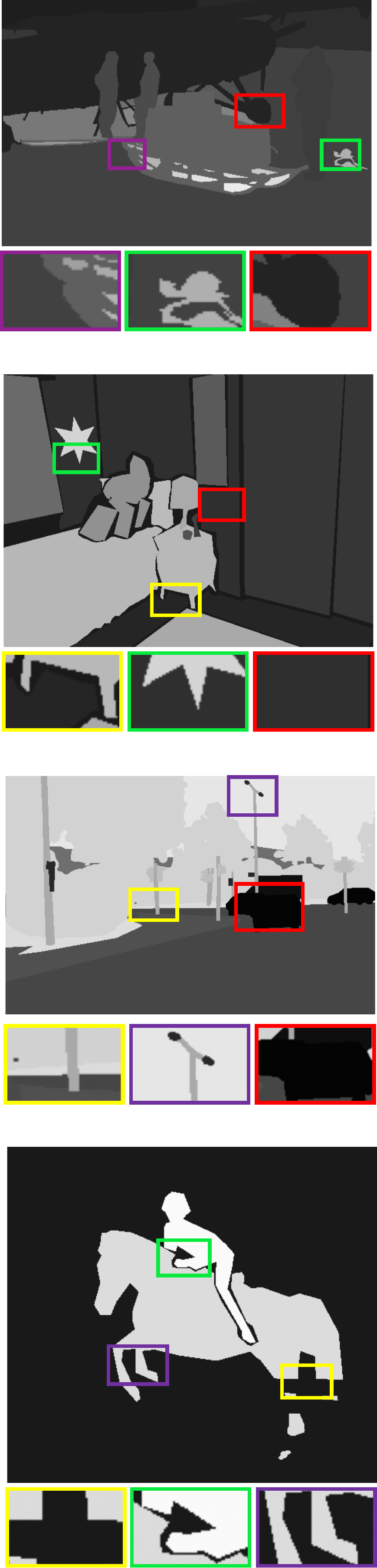}
    \caption{Label}
  \end{subfigure}
  \hfill
  \begin{subfigure}[b]{0.11\textwidth}
    \includegraphics[width=\textwidth]{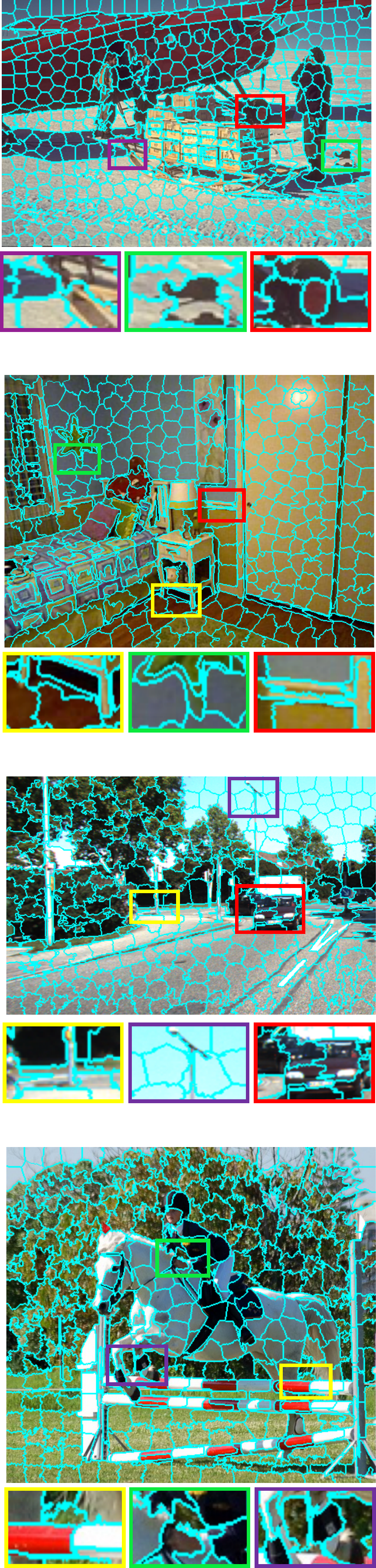}
    \caption{SLIC}
  \end{subfigure}
  \hfill
  \begin{subfigure}[b]{0.11\textwidth}
    \includegraphics[width=\textwidth]{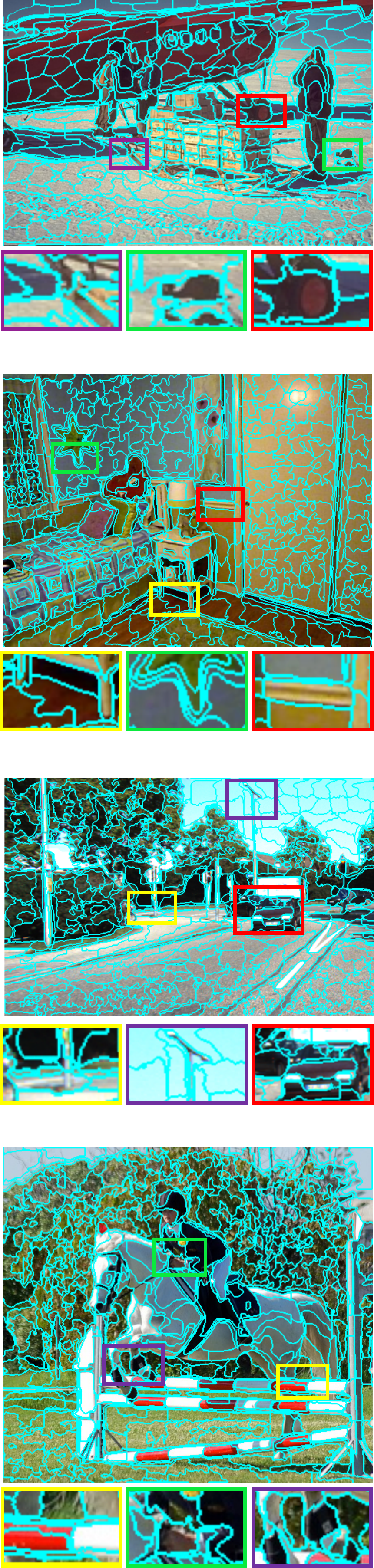}
    \caption{SSN}
  \end{subfigure}
  \hfill
  \begin{subfigure}[b]{0.11\textwidth}
    \includegraphics[width=\textwidth]{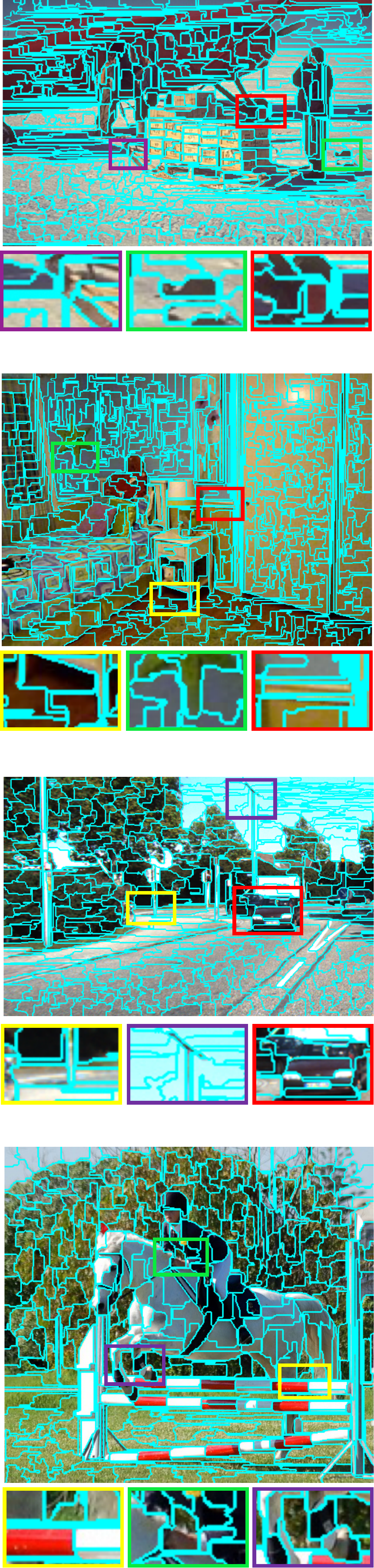}
    \caption{SEAL}
  \end{subfigure}
  \hfill
  \begin{subfigure}[b]{0.11\textwidth}
    \includegraphics[width=\textwidth]{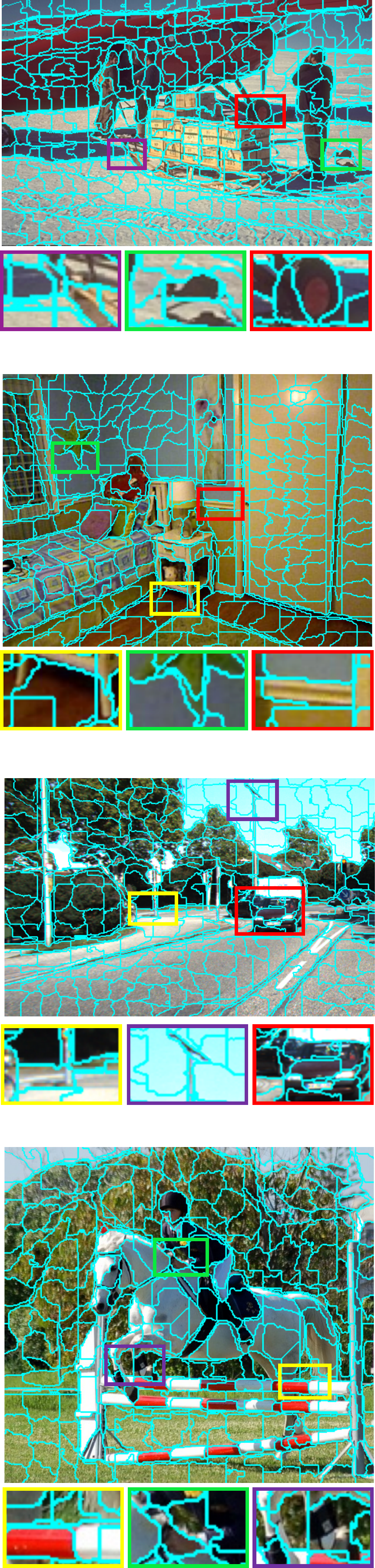}
    \caption{SCN}
  \end{subfigure}
  \hfill
  \begin{subfigure}[b]{0.11\textwidth}
    \includegraphics[width=\textwidth]{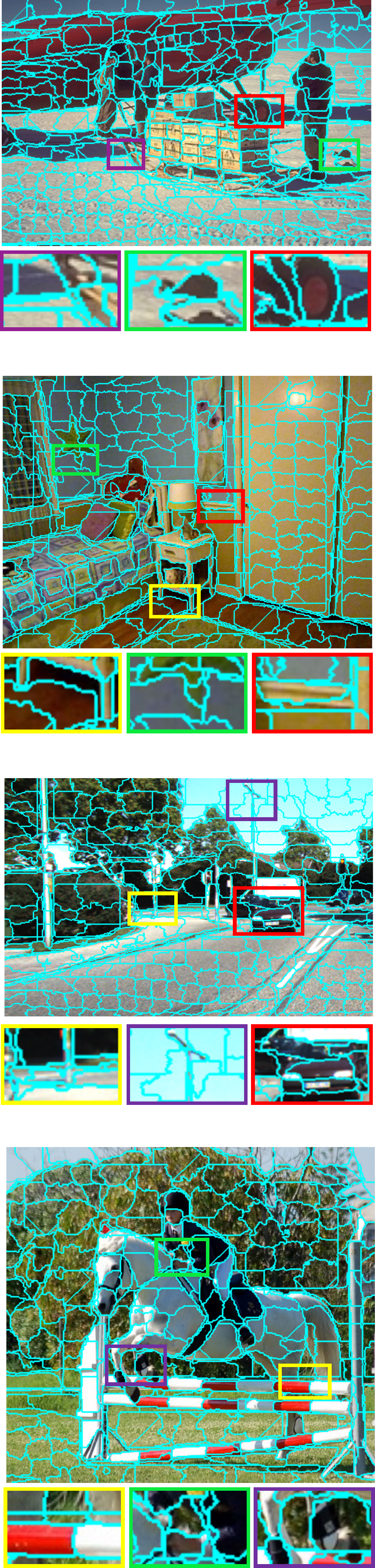}
    \caption{AINet}
  \end{subfigure}
  \hfill
  \begin{subfigure}[b]{0.11\textwidth}
    \includegraphics[width=\textwidth]{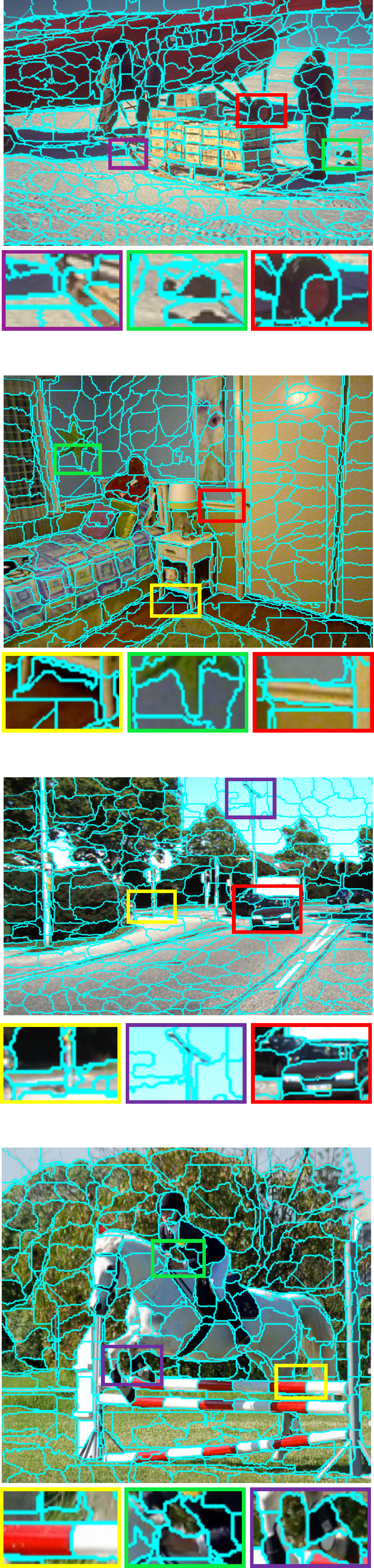}
    \caption{Ours}
  \end{subfigure}
  
  \caption{Visualization results of ours and previous methods. Compared to other superpixel algorithms, our method achieves better boundary adherence when facing unseen images. (From Top to Bottom: examples from BSDS, NYUv2, KITTI, VOC2012.)}
  \label{fig:visual}
\end{figure*}

\noindent\textbf{Qualitative comparison.} Fig.\ref{fig:visual} shows the qualitative results of six state-of-the art methods on dataset BSDS, NYUv2, KITTI, and Pascal VOC. Our method has better performance when facing critical object contours. Please see more visual results in the supplementary.

\begin{figure}[t]
  \centering
    \includegraphics[width=0.95\columnwidth]{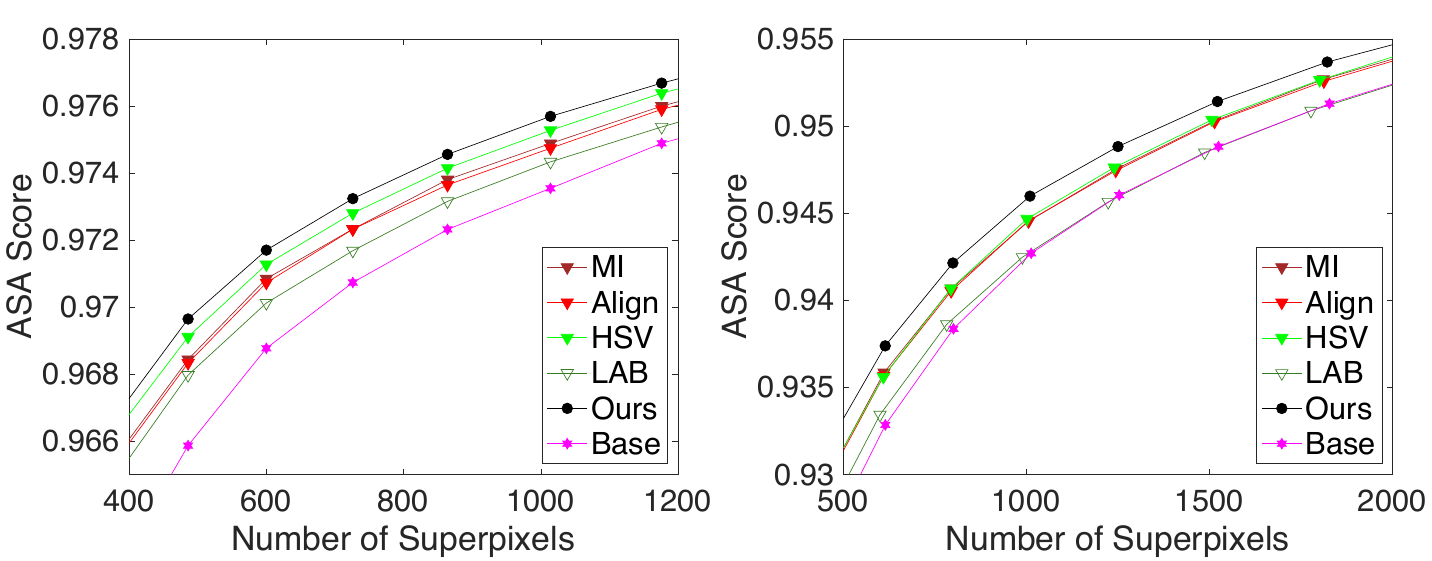} 
  \caption{Component analysis. From left to right: ASA score on the BSDS and NYU datasets.}
  \label{fig:able1}
\end{figure}

\begin{figure}[t]
  \centering
    \includegraphics[width=0.95\columnwidth]{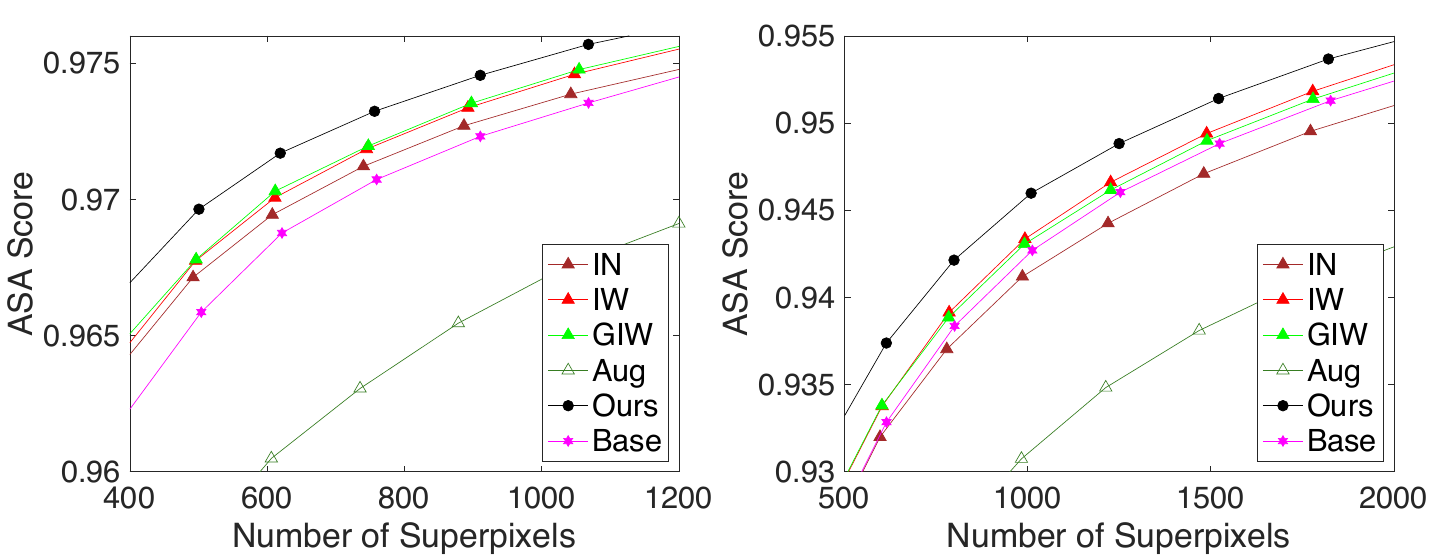} 
  \caption{Comparison with other style removal methods. From left to right: ASA score on the BSDS and NYU datasets.}
  \label{fig:able2}
\end{figure}

\subsection{Ablation Study}
\label{able}

We conduct experiments primarily to address the following key questions: 

\begin{itemize}
    \item \textbf{Q1}: The choices of auxiliary modalities, and the difference from using them as a form of data augmentation. 
    \item \textbf{Q2}: Since our goal is to extract shared information from image pairs with significant style differences, why not use alignment directly. 
    \item \textbf{Q3}: Can other universal style-removal methods improve the performance of the CDS model structure.
\end{itemize}

\noindent\textbf{Component Analysis.} As illustrated in Fig.\ref{fig:able1}, we first study the choice of auxiliary modality, i.e., HSV, LAB, and gradient map (Ours), for question \textbf{Q1}. The performance ranking is Ours $>$ HSV $>$ LAB. As expected, the dissimilarity between the auxiliary modality and the RGB modality is positively correlated with the experimental performance. The LAB color space retains two color channels, while the image gradient map does not possess the same color information as RGB. Then, to answer the question \textbf{Q2}, the conducted experiments show that employing either feature alignment constraints or style mutual information minimization alone reduces the performance of the method, but both are still superior to the baseline method\cite{yang2020superpixel-FCN}. We consider that (1) When only using alignment constraints, it actually does not decouple and remove the attribute noise of the dataset, but forcibly brings the two modalities closer in the feature space, resulting in the retention of some irrelevant information for superpixel segmentation. (2) Since the two modalities share the parameters of the superpixel generator, it is equivalent to implicit alignment. However, without the pixel correlation guidance for shared information extraction, it leads to a decrease in performance.

\noindent\textbf{Comparison with universal style-removal approaches.} We compare our method to other style-removal manners in Fig.\ref{fig:able2} to further verify the our effectiveness. Firstly, to supplement the question \textbf{Q1}, we employ the auxiliary modality transform as a data augmentation strategy (model Aug), i.e, applying the random modal transition by probability 0.3 for training the single-modality model. Due to the significant style differences between modalities and the original RGB images, it confuses the lightweight feature extractor of the superpixel algorithm, resulting in a severe performance decline. Then we compare our methods with the IN \cite{ulyanov2017IN}, IW \cite{li2017whiten}, GIW \cite{cho2019GIW} for question \textbf{Q3}. As shown in the figure, our method achieve the best performance. Specifically, the generalization performance of Instance Normalization (IN) falls behind our baseline. We believe this is because during the training phase, Instance Normalization introduces additional errors due to significant individual variations.

\subsection{Application}
\label{app}
Superpixel algorithms provide better feature representation for downstream tasks, improving model performance and efficiency. The performance of superpixels in downstream tasks also demonstrates the effectiveness of superpixel algorithms. Table.\ref{tab:app} reports the performance comparison on downstream semantic segmentation tasks. We leverage the pretrained Deeplab\cite{chen2014deeplab} and Bilateral Inception (BI) module\cite{gadde2016BI} and directly replace the superpixel segmentation generated by gSLICr\cite{gSLICr_2015} with different superpixel results. The first row indicates the official implementation with 1000 superpixels. The following four lines show the results using 600 superpixels. Our method achieved an mIOU of 79.02 on the reduced validated set used by \cite{gadde2016BI} of Pascal VOC2012, outperforming other superpixel algorithms.  

\begin{table}
  \centering
  \resizebox{0.6\columnwidth}{!}{%
  \begin{tabular}{l|l|c}
  \hline
  \hline
    Base Modal & Methods  & IoU \\
    \hline
    
     \multirow{5}{*}{DeepLab} & BI (gSLICr) \quad & 78.54  \\
      & BI (ETPS)  \quad & 77.67  \\
      & BI (SCN)  \quad & 78.90  \\
      & BI (AINet)  \quad & 78.96  \\
      & BI (Ours)  \quad & 79.02  \\
\hline
\hline
  \end{tabular}%
  }
  \caption{Superpixel algorithms with downstream segmentation. IoU comparison on the Pascal VOC dataset.}
  \label{tab:app}

\end{table}

\section{Conclusion}
In this paper, we propose the Content Disentangle Superpixel algorithm to eliminate the dataset style noise that exists in the learnable superpixel features and reduce the feature-level distribution difference between training data and open-world data for superpixel segmentation. Unlike other deep superpixel algorithms that use single-modal training, we introduce auxiliary modalities to assist in decoupling the RGB image features into invariant image content and style noise. Specifically, we propose local-grid correlation alignment to maximize the selected image content information across modalities and learn the invariant inter-pixel correlations for superpixel generation. Then, we propose the global-style mutual information minimization to the minimize the upper bound of mutual information for style information, and prevent the degenerate solution during the disentangle process. Experimental results demonstrate that our method effectively mitigates the impact of style noise on deep superpixel algorithms and achieves superior results compared to existing methods on four different datasets.

\section{Acknowledgements}
This work was supported in part by the National Key R\&D Program of China (No.2021ZD0112100), and National Natural Science Foundation of China (No.61972022, No.U1936212, No.62120106009, No.52202486).

\clearpage
\appendix
\subsection{Appendix.}

\subsubsection{Alignment and Mutual Information.}
In this subsection, we provide a theoretical proof to validate that, while keeping other conditions constant, local-grid correlation alignment can increase the mutual information between features. Define the features before alignment of image $\mathcal{I}$ and auxiliary modal $\mathcal{A}$ are $\mathcal{F}_{\mathcal{I}}$ and $\mathcal{F}_{\mathcal{A}}$, while features after alignment are  $\mathcal{F}^{\prime}_{\mathcal{I}}$ and $\mathcal{F}^{\prime}_{\mathcal{A}}$. Based on the definition of mutual information:
\begin{equation}
\begin{split}
    \mathrm{MI}(\mathcal{F}_{\mathcal{I}},\mathcal{F}_{\mathcal{A}}) = H(\mathcal{F}_{\mathcal{I}}) - H(\mathcal{F}_{\mathcal{I}}|\mathcal{F}_{\mathcal{A}})\\
    \mathrm{MI}(\mathcal{F}^{\prime}_{\mathcal{I}},\mathcal{F}^{\prime}_{\mathcal{A}}) = H(\mathcal{F}^{\prime}_{\mathcal{I}}) - H(\mathcal{F}^{\prime}_{\mathcal{I}}|\mathcal{F}^{\prime}_{\mathcal{A}})
\end{split}
\end{equation}
To show that $\mathrm{MI}(\mathcal{F}^{\prime}_{\mathcal{I}},\mathcal{F}^{\prime}_{\mathcal{A}})$ is larger than $\mathrm{MI}(\mathcal{F}_{\mathcal{I}},\mathcal{F}_{\mathcal{A}})$, we calculate the gap $\Delta$ between them:

\begin{equation}
\begin{aligned}
    \Delta &= \mathrm{MI}(\mathcal{F}^{\prime}_{\mathcal{I}},\mathcal{F}^{\prime}_{\mathcal{A}}) -\mathrm{MI}(\mathcal{F}_{\mathcal{I}},\mathcal{F}_{\mathcal{A}}) \\
    &= (H(\mathcal{F}^{\prime}_{\mathcal{I}}) - H(\mathcal{F}_{\mathcal{I}})) + (H(\mathcal{F}_{\mathcal{I}}|\mathcal{F}_{\mathcal{A}}) - H(\mathcal{F}^{\prime}_{\mathcal{I}}|\mathcal{F}^{\prime}_{\mathcal{A}}))
\end{aligned}
\end{equation}
With other conditions held constant, before and after performing alignment: (1) The feature $\mathcal{F}$ is strongly constrained by the same label and superpixel segmentation loss. (2) $\mathcal{F}$ describes the spatial-wise inter-pixel correlations. As a result, the entropy of $\mathcal{F}$ itself does not undergo significant changes before and after alignment, i.e., $H(\mathcal{F}^{\prime}_{\mathcal{I}}) - H(\mathcal{F}_{\mathcal{I}}) \approx 0$. Then,
\begin{equation}
\begin{aligned}
    \Delta &\approx (H(\mathcal{F}_{\mathcal{I}}|\mathcal{F}_{\mathcal{A}}) - H(\mathcal{F}^{\prime}_{\mathcal{I}}|\mathcal{F}^{\prime}_{\mathcal{A}}))\\
    &= -\mathcal{P}(\mathcal{F}_{\mathcal{I}}|\mathcal{F}_{\mathcal{A}}) \cdot \log \mathcal{P}(\mathcal{F}_{\mathcal{I}}|\mathcal{F}_{\mathcal{A}}) \\
    & \quad\quad+ \mathcal{P}(\mathcal{F}^{\prime}_{\mathcal{I}}|\mathcal{F}^{\prime}_{\mathcal{A}}) \cdot \log \mathcal{P}(\mathcal{F}^{\prime}_{\mathcal{I}}|\mathcal{F}^{\prime}_{\mathcal{A}}) > 0.
\end{aligned}
\end{equation}

\subsubsection{Training \& Inference.}
The Content Disentangle Superpixel algorithm is trained in an end-to-end manner. The auxiliary modality is used only during the training phase. Due to the modality barrier, we employ separate $\Psi$ to extract features from the two modalities. Then, a shared-parameter content selective gate and superpixel decoder are utilized to decouple superpixel-friendly content and predict the association map $\mathcal{Q}$. In this manner, we can easily remove the auxiliary modalities during the inference phase. Moreover, the variational distribution network $h^{\theta}$ is optimized with independent adam optimizer with learning rate 5e-3.

During inference, since the $Q$ indicates the probability that each pixel is attributed to its nine nearby superpixel grids. The size $d$ of initial superpixel grid should remain constant both during the training and inference phases. Here, $d = 16$ is controlled by the $\mathcal{L}_{sp}$ \cite{yang2020superpixel-FCN}. As a result, we need to resize the images to an appropriate resolution to control the number of generated superpixels, e.g., if we want to generate $({sp}_h, {sp}_w)$ for an image, we need to resize the image into $({sp}_h \times d, {sp}_w \times d)$. After calculate the superpixel assignment, the output is "nearest" resized to the original resolution.

\subsection{Experimental Details}
\subsubsection{Datasets and License} We conduct experiments mainly on five datasets: BSDS500 \cite{bsds500}, NYUv2 \cite{nyuv2}, KITTI \cite{geiger2012kitti}, Pascal VOC \cite{everingham2015voc}. According to \cite{yang2020superpixel-FCN,wang2021ainet}'s protocols, we trained our model on BSDS \cite{bsds500} dataset only and perform evaluation on different datasets. BSDS dataset contains at least five segmentation results $Y_{seg}=\{y_1,y_2,...y_n\}$ from different annotators for each image $I$. We divide the dataset into different individual samples by segmentation labels (e.g., $I\rightarrow y_1, I\rightarrow y_2, ... , I\rightarrow y_n$). Thus, each label is involved in the computation during the training and evaluation phases for fair comparison. Finally, we transform the training and validation sets of BSDS500 into 1633 samples for training and transform the test set into 1063 test samples for evaluation. Among them, KITTI dataset uses the CC BY-NC-SA 3.0 license, while other datasets are open source but do not provide specific licenses. 

\begin{figure*}[t]

\centering
\includegraphics[width=\textwidth]{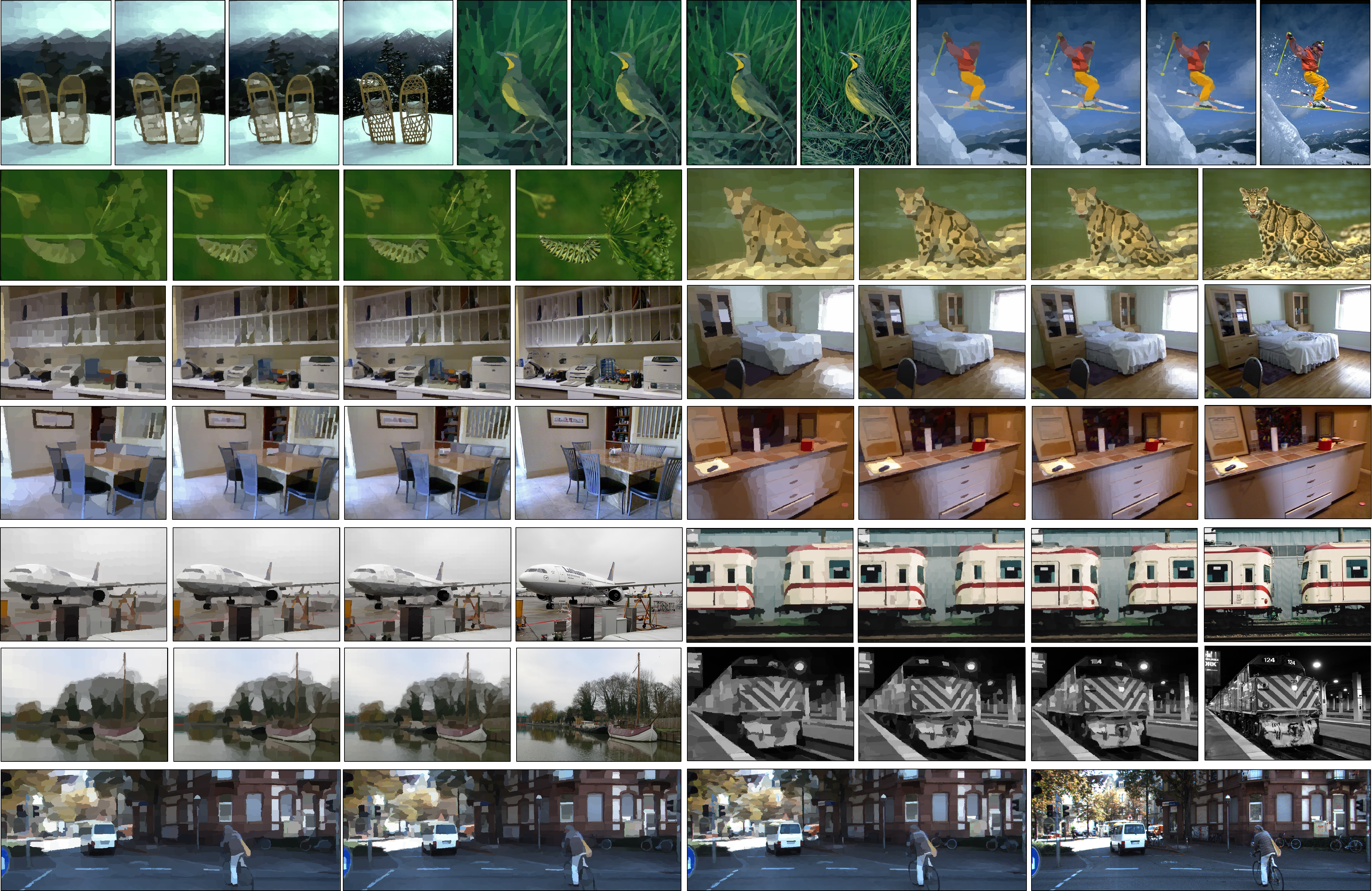}
\caption{Reconstruction Results. We fill superpixels with average value of contained pixels. Each group of images contains three Superpixel resolutions and original image. Our method can recover accurate target contours with fewer superpixels.} 
\label{fig:recon}
\end{figure*}

\subsubsection{Metrics}

In the main manuscript, we use the Achievable Segmentation Accuracy (ASA)\cite{liu2011ERS}, Boundary Recall-Precision Curve (BR-BP), and Undersegmentation Error (UE), which are most widely-used to measure superpixel segmentation qualities as our primary metrics. Formally, let $\mathcal{S}={\{S_i\}}_{i=1}^K$ and $\mathcal{G}=\{G_j\}$ be the partitions of superpixel segmentation and groundtruth segmentation, respectively. Then, the ASA is defined as:
\begin{equation} \label{ASA}
    \operatorname{ASA}(\mathcal{S})=\frac{1}{N} \sum_{i} \max _{j}\left\{\left|S_{i} \cap G_{j}\right|\right\}.
\end{equation}
The ASA indicates the achievable segmentation accuracy when each superpixel category is correctly labeled. However, as shown in Eq.\ref{ASA}, the ASA is concerned with the overlap area, which results in maintaining more than $92\%$ results even if each superpixel does not achieve reliable inter-class similarity. Typically, the superpixel algorithm obtains an ASA within $[92\%, 100\%]$. To better distinguish performance differences, we introduce another popular superpixel metric, BR-BP, to asses boundary adherence given ground truth. BR and BP are defined as:
\begin{equation} \label{BR}
\begin{aligned}
    \mathrm{BR}(\mathcal{S})&=\frac{\mathrm{TP}(\mathcal{G}, \mathcal{S})}{\mathrm{TP}(\mathcal{G}, \mathcal{S})+\mathrm{FN}(\mathcal{G}, \mathcal{S})},\\
\mathrm{BP}(\mathcal{S})&=\frac{\mathrm{TP}(\mathcal{G}, \mathcal{S})}{\mathrm{TP}(\mathcal{G}, \mathcal{S})+\mathrm{FP}(\mathcal{G}, \mathcal{S})}, 
\end{aligned}
\end{equation} 
where $\mathrm{FN}(\mathcal{G}, \mathcal{S})$ and $\mathrm{TP}(\mathcal{G}, \mathcal{S})$ are the number of false negative and true positive boundary pixels in $\mathcal{S}$ with respect to $\mathcal{G}$. And $\mathrm{FP}$ is the number of false positive pixels. Then, UE is another primary metric used as the main evaluation metric by multiple superpixel related works\cite{levinshtein2009turbopixels,stutz2018superpixels,achanta2012slic}. Given a region from segmentation ground truth $G_j$ and a set of superpixels required to cover it, i.e., $S_i | (S_{i} \cap G_{j} \neq \emptyset)$, Under-segmentation error measures how many pixels from $S_i$ “leak” across the boundary of $G_j$. Overall, UE score has three versions\cite{levinshtein2009turbopixels,achanta2012slic,neubert2012ue3}, and we use the latest formulations proposed by \cite{neubert2012ue3}:
\begin{equation} \label{UE}
\mathrm{UE}(\mathcal{S})=\frac{1}{N} \sum_{j} \sum_{S_{i} \cap G_{j} \neq \emptyset} \min \left\{\left|S_{i} \cap G_{j}\right|,\left|S_{i}-G_{j}\right|\right\}
\end{equation}
UE is more concerned with the existence of different classes within each superpixel cluster compared to the achievable segmentation accuracy.

\subsubsection{Compactness Analysis.} 
\begin{figure}[t]
  \centering
    \includegraphics[width=0.95\columnwidth]{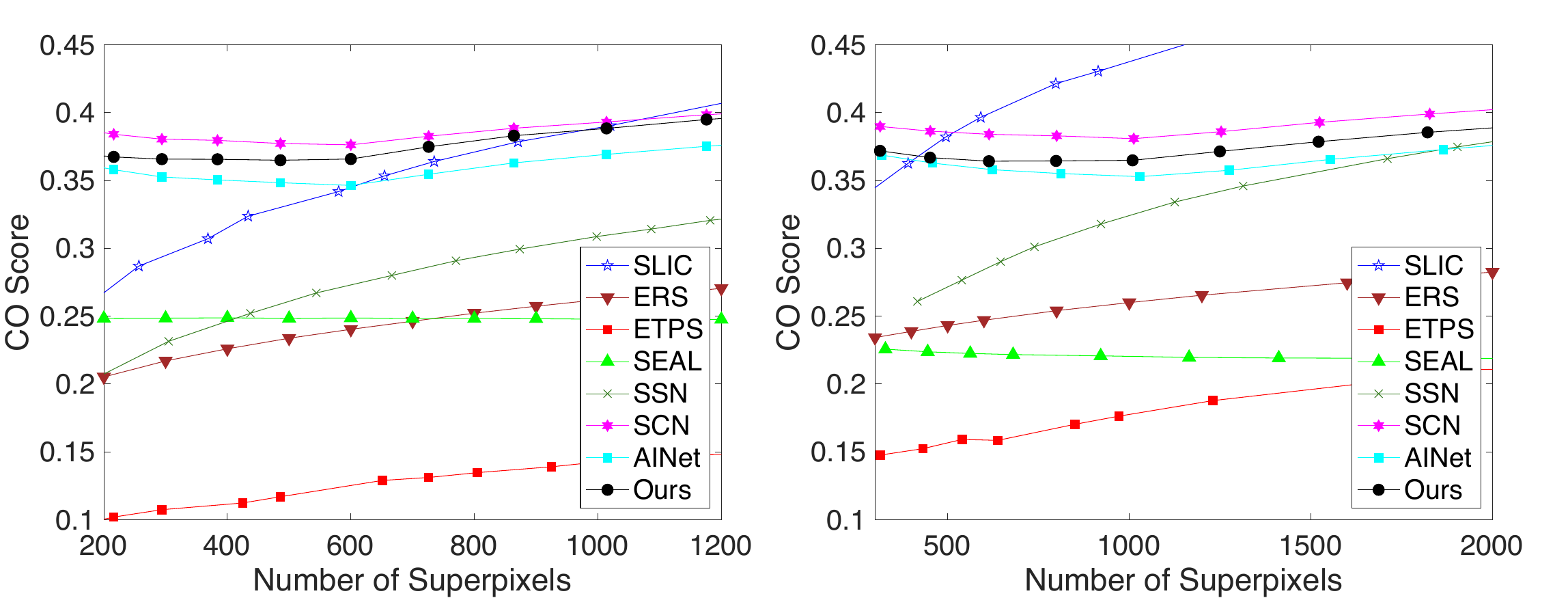} 
  \caption{Compactness Comparison. From left to right: BSDS and NYU datasets.}
  \label{fig:supco}
\end{figure}

Compactness (CO) \cite{achanta2012slic}  metric reflects the regularity of superpixels:
\begin{equation}
\mathrm{CO}(\mathcal{S})=\frac{1}{N} \sum_{S_{i}}\left|S_{i}\right| \frac{4 \pi A\left(S_{i}\right)}{P\left(S_{i}\right)}
\end{equation}
CO score compares the area $A\left(S_{i}\right)$ of each superpixel $S_i$ with the area of a circle with same perimeter $P\left(S_{i}\right)$. In general, CO is related to the superpixel initialization seed point or grid density, and also affected by the regularity of the target object in the image.  Since CO cannot directly reflect the accuracy of superpixel clustering, for example, regular square grids often yield high CO scores and are consequently employed as \textbf{secondary} metrics to assess superpixel performance. Fig.\ref{fig:supco} reports the CO score on two datasets and we can find that:
\begin{itemize}
    \item Since we define superpixels as a neighborhood classification problem, our method, like SCN \cite{yang2020superpixel-FCN} and AINet \cite{wang2021ainet}, generates superpixels with high compactness.
    \item The CO metric of our methods is slightly lower than that of the SCN and SLIC methods. In fact, this is a normal phenomenon, because better contour fitting ability often comes at the cost of reduced regularity of superpixels. Overall, the improvement of our algorithm in major metrics such as ASA is more valuable.
\end{itemize}

\subsubsection{Ablation Study}

In this subsection, we provide an explanation of several specific model settings that we used in the ablation study.  (a) Model \textit{Aug}: we incorporate assistant modalities (HSV, LAB, gradient map) as data augmentation. With a probability of 0.3, randomly selected one modality to replace the RGB image with the same probability for each modality. (b) Model \textit{IN}: replace all batch normalization layers with instance normalization in the network. (c) Model \textit{IW}: perform whitening loss only on the features after feature extraction. (d) Model \textit{GIW}: perform group instance whitening only on the features after feature extraction. Except for the model modifications mentioned above, all hyperparameters used for model training are kept consistent with our main experiment.

\subsection{Visualization}

\subsubsection{Reconstruction visualization.} To demonstrate the quality of image representation by superpixels, we replace the pixel values within each superpixel with the average value of the entire superpixel. As shown in the Fig.\ref{fig:recon}, even with a smaller number of superpixels, our method still maintains clear object boundaries.

\subsubsection{Demo.} We have provided a simple-to-reproduce demo with our code, which can visualize the superpixel segmentation results of any image.

\subsection{More Discussion}
\subsubsection{Social Impact.} Superpixels can be utilized in a wide range of visual tasks. Existing deep superpixel algorithms, as low-level image representation methods, have overlooked the issue of model dependency on dataset distribution induced by training, thereby falling short of meeting the practical demands of applying superpixels in open data environments. To tackle this challenge, we introduce assistant modalities and present the first superpixel algorithm that effectively addresses this problem. As a result, our approach CDS has better advantages in some scenarios, such as aiding manual pixel-level semantic annotation of unseen data.
\subsubsection{Limitation.} In addition to the boundary adherence, computational efficiency is also very important indicator for superpixel segmentation. Although our method is relatively computationally efficient, the inference consumes additional computational resources when generating a larger number of superpixels. This is also a common issue for all methods that define superpixels as a neighborhood classification problem.

\bibliography{aaai24}

\end{document}